\documentclass{article}

\usepackage[dblblindworkshop, final]{neurips_2025}

\workshoptitle{Frontiers in Probabilistic Inference: Sampling Meets Learning}

\usepackage[utf8]{inputenc} 
\usepackage[T1]{fontenc}    
\usepackage{hyperref}       
\usepackage{url}            
\usepackage{booktabs}       
\usepackage{amsfonts}       
\usepackage{nicefrac}       
\usepackage{microtype}      
\usepackage{xcolor}         
\usepackage{graphicx}   
\usepackage{subcaption} 
\usepackage{amsmath}
\usepackage{amsthm}
\usepackage{pifont}
\usepackage[ruled,vlined]{algorithm2e}

\newtheorem{proposition}{Proposition}

\title{Sampling from Energy distributions \\ with Target Concrete Score Identity}

\author{%
  Sergei Kholkin\\
  Applied AI Institute, Moscow, Russia \\
  \texttt{kholkinsd@gmail.com} \\
  \And
  Francisco Vargas \\
  Xaira Therapeutics \\
  \texttt{vargfran@gmail.com} \\
  \And
  Alexander Korotin \\
  Applied AI Institute, Moscow, Russia \\
  \texttt{iamalexkorotin@gmail.com} \\
}

\begin{document}

\maketitle

\begin{abstract}

  We introduce the \textit{Target Concrete Score Identity Sampler} (TCSIS), a method for sampling from unnormalized densities on discrete state spaces by learning the reverse dynamics of a Continuous-Time Markov Chain (CTMC). Our approach builds on a forward in time CTMC with a uniform noising kernel and relies on the proposed \textit{Target Concrete Score Identity}, which relates the concrete score, the ratio of marginal probabilities of two states, to a ratio of expectations of Boltzmann factors under the forward uniform diffusion kernel. This formulation enables Monte Carlo estimation of the concrete score without requiring samples from the target distribution or computation of the partition function. We approximate the concrete score with a neural network and propose two algorithms: \textit{Self-Normalized TCSIS} and \textit{Unbiased TCSIS}. Finally, we demonstrate the effectiveness of TCSIS on problems from statistical physics.
\end{abstract}

\vspace{-3mm}
\section{Introduction}
\vspace{-3mm}






Let $p(x)$ be a probability distribution on finite space $S$:
\vspace{-1mm}
\begin{equation}
    p(x) = \frac{\overline{p}(x)}{Z}, \quad Z = \sum_{x \in S} \overline{p}(x),
\end{equation}
\vspace{-5mm}

where $\overline{p}(x): S \rightarrow \mathbb{R}_+$ is the unnormalized density, also known as Boltzman distribution, and the normalization constant $Z$ is generally intractable. In this work we are interested in sampling from $p(x)$ given only $\overline{p}(x)$. This task is central in Bayesian inference \cite{murray2006mcmc}, statistical physics \cite{newman1999monte}, and computational biology \cite{lartillot2004bayesian}, yet remains challenging in large state spaces.



Sampling from unnormalized densities is well studied in continuous spaces, where Markov Chain Monte Carlo (MCMC) serves as the standard baseline \cite{neal2011mcmc, metropolis1953equation}. However, MCMC methods often suffer from poor mixing, slow convergence, and local trapping \cite{neal2011mcmc}. These challenges have motivated the development of neural samplers \cite{vargasdenoising, vargastransport_CMCD, havensadjoint}, parameterized by neural networks to improve efficiency.

In discrete state spaces, MCMC methods are also a strong baseline \cite{lireheated_DMALA, grathwohl2021oops_GWG}, but neural approaches are only beginning to emerge. Recent work on discrete diffusion models and learned CTMCs \cite{campbell2022continuous, lou2024discrete, sahoo2024simple_MDLM} has enabled neural samplers for combinatorial optimization \cite{sanokowskiscalable_SDDS} and statistical physics \cite{holderriethleaps, ou2025dnfs}. For example, SDDS \cite{sanokowskiscalable_SDDS} trains a diffusion via policy gradients and self-normalized importance sampling, while LEAPS \cite{holderriethleaps} and DNFS \cite{ou2025dnfs} use temperature annealing between a Boltzmann distribution and noise, combined with coordinate descent to learn a CTMC sampler.

The goal of this work is to leverage recent advances in discrete diffusion models and learnable CTMCs, specifically probability factorization and concrete scores, to design a learnable sampler from unnormalized densities. We state a \textit{Target Concrete Score Identity} that allows to estimate the concrete score at given time-moment $t$ for uniform discrete diffusion forward process \cite{austin2021structured, schiffsimple_UDLM} given only the energy function. Further we propose \textit{Target Concrete Score Identity Sampler} (\textbf{TCSIS}) framework and two practical algorithms: 1) self normalized, to learn the concrete score directly 2) unbiased to learn the unnormalized marginal densities and construct concrete score out of them during the inference stage. Finally we do evaluate our method on the problems from statistical physics.

\vspace{-3mm}
\section{Continuous-Time Markov Chains}
\vspace{-2mm}




We begin by introducing Continuous-Time Markov Chains (CTMCs) \cite{campbell2022continuous, lou2024discrete}.  We consider finite state spaces $S$ of the form $S = \mathcal{X}^d$, where $\mathcal{X} \in \{1,..,V\}$ is a set of tokens. robability distributions over the discrete space $S$ are represented by probability mass vectors $p \in \mathbb{R}^{|S|}$, where each entry $p(x)$ denotes the probability of state $x$, and the entries sum up to one. A CTMC defines a family of time-indexed distributions $p_t$ evolving according to linear ordinary differential equation:



\vspace{-5mm}
\begin{equation} \label{eq:CTMC_Kolmogorov_forward}
    \frac{\partial p_t}{\partial t}=Q_tp_t, \quad p_{t=0}(x)=p_0(x),
\end{equation}
\vspace{-3mm}

where $Q_t(y, x): S \times S \rightarrow \mathbb{R}$ is a transition rate matrix which captures the rate of change of transition probabilities and have non-negative non-diagonal entries and columns which sum to zero.
Since the ODE (\ref{eq:CTMC_Kolmogorov_forward}) describes the infinitesimal dynamics of a CTMC, the process can be simulated by discretizing time into small steps. Over an infinitesimal interval, the chain either stays in its current state or transitions to another, with rates governed by the entries of the  transition rate matrix $Q_t$.






Finally, this process has a well-known time-reversal \cite{campbell2022continuous, lou2024discrete}, characterized by another matrix $\overline{Q}_t$:

\vspace{-3mm}
\begin{equation}
    \frac{\partial p_{T-t}}{\partial t}=\overline{Q}_{T-t}p_{T-t}, \quad 
    \overline{Q}_t(y, x) = \left\{
    \begin{array}{l}
    \frac{p_t(y)}{p_t(x)}Q_t(x, y), y \neq x \\
    -\sum_{y\neq x}Q_t(x, y), y=x
    \end{array}
    \right.
\end{equation}
\vspace{-2mm}


Here, ratio $\frac{p_t(y)}{p_t(x)} = s_t(x)_y$ is known as the \textbf{concrete score} \cite{meng2022concrete, lou2024discrete}. In literature, it is typically represented in vector form, $s_t(x) = \left[\frac{p_t(y)}{p_t(x)}\right]_{y \in \mathcal{X}^d}$, so that $s_t(x): \mathcal{X}^d \times \mathbb{R} \rightarrow \mathbb{R}^{|\mathcal{X}^d|}$. Obtaining the concrete score is sufficient for defining the reverse dynamics.






We call by \textit{discrete diffusion model} a special case of a CTMC in which the forward process, defined by $Q_t$, gradually corrupts the data toward a simple distribution. The forward dynamics is known, while the backward process is learned, typically from data. To remain scalable, such models parameterize transition rate matrices or concrete scores in a factorized manner across dimensions \cite{campbell2022continuous, meng2022concrete}, which is equivalent to modeling transitions between sequences $y, x \in \mathcal{X}^d$ restricted to their 1-Hamming neighborhoods. In that case the concrete score would change notion $s_t(x): \mathcal{X}^d \times \mathbb{R} \rightarrow \mathbb{R}^{d \times V}$. For additional information see Appendix~\ref{appx:factorization}. For simplicity we use general notation, but in practice we adopt this factorization as well.


We focus on uniform discrete diffusion models \cite{schiffsimple_UDLM, lou2024discrete, austin2021structured}, where the forward dynamics is defined via uniform noising kernel and is constructed so that the terminal distribution is uniform categorical noise. For additional information see Appendix~\ref{appx:uniform_diff}.








\vspace{-3mm}
\section{Method}
\vspace{-2mm}

In the Section~\ref{sec:TCSI} we provide an expression for the concrete score of discrete diffusion with uniform noising process. In Section~\ref{sec:TCSIS} we propose general framework for constructing a sampler from unnormalized density $\overline{p}(x) = p(x)Z$ and a way to learn related concrete scores without access to samples from $p(x)$. In Sections \ref{sec:TCSIS_snis} and \ref{sec:TCSIS_unbiased} we propose two implementations of described in Section~\ref{sec:TCSIS} framework with different loss functions and different parametrizations of a neural network.

\vspace{-3mm}
\subsection{Target Concrete Score Identity} \label{sec:TCSI}
\vspace{-2mm}

The central component of our work is the notion of the \textbf{concrete score} \cite{meng2022concrete}. Prior works \cite{meng2022concrete, lou2024discrete} propose methodologies for learning the concrete score from data samples. 




\begin{proposition}[Target Concrete Score Identity]\label{th:TCS_Identity}
    Consider a probability distribution $p(x)$ with finite support $S=\mathcal{X}^d$, where $\mathcal{X}=\{1, ..., V\}$ and its unnormalized density $\overline{p}(x) = p(x)Z$, where $Z$ is unknown. Consider the continuous in time dynamics determined by uniform forward noising kernel (\ref{eq:uniform_forward}). Then the ratio of its marginal distributions at time $t$, i.e, concrete score $s_t(x)$, has the form:

\vspace{-3mm}
    $$s_t(x)_y = \frac{p_t(y)}{p_t(x)} = \frac{\mathbb{E}_{p_{t|0}(x_0|y)}[\overline{p}(x_0)]}{\mathbb{E}_{p_{t|0}(x_0|x)}[\overline{p}(x_0)]}.$$

\end{proposition}

For proof see Appendix~\ref{appx:proofs}. This proposition can be viewed as discrete state space counterpart of Proposition 2.1 in \cite{de2024target_TSM}. Further, Proposition~\ref{th:TCS_Identity} enables Monte Carlo estimation of concrete score $s_t(x)_y$, given $x$ and $y$. Relatedly, \cite{zhang_target_thornton} introduces a similar to Proposition~\ref{th:TCS_Identity} identity for the concrete score of the posterior $p_{0|t}$, which they employ for training of Masked Diffusion Models \cite{sahoo2024simple_MDLM}.

\vspace{-3mm}
\subsection{Target Concrete Score Identity Sampler}\label{sec:TCSIS}
\vspace{-2mm}



Next, we use the Target Concrete Score Identity to learn a reverse discrete diffusion for sampling from an unnormalized density $\overline{p}(x) = p(x)Z$. Following \cite{meng2022concrete}, we approximate the concrete score with a neural network $s_t(x)\sim s_{t}^\theta(x)$ and train it by minimizing a loss. Since Proposition~\ref{th:TCS_Identity} gives a direct expression for the concrete score, it suffices to minimize the discrepancy between $s_t(x_t)$ and $s_{t}^\theta(x_t)$ over all $x_t$. We therefore propose the following objective:



\vspace{-3mm}
\begin{equation}\label{eq:TCSIS_base}
    L_\theta = \mathbb{E}_{q(x_t), t}[D(s_t(x_t), s_{t}^\theta(x_t))],
\end{equation}

\vspace{-2mm}
where $q(x_t)$ is \textit{arbitrary distribution with full support}, $t \sim U[0, T]$ and $D$ is divergence applicable for concrete scores. The neural network parametrization of the concrete score is given by $s_{t}^\theta(x_t): \mathcal{X}^d \times \mathbb{R} \rightarrow \mathbb{R}^{d \times V}$. We call this framework \textit{Target Concrete Score Identity Sampler} (\textbf{TCSIS}) and the learning procedure is described in Algorithm~\ref{alg:TCSIS_train}. 
We note that our training procedure in general is \textit{is simulation free}. After the concrete score $s_{\theta, t}$ is learned the sampling procedure follows regular discrete diffusion models practice and is given in Section 4 \cite{lou2024discrete}.



As one can notice the distribution $q(x_t)$ \textit{can be varied as long as it has full support}. Following Proposition 1 \cite{akhound2024iterated_iDEM} we suppose that more informative $q(x_t)$ that closely follows ground truth $p_t(x_t)$ is beneficial. We parametrize $q(x_t)=\mathbb{E}[q(x_0)p_{t|0}(x_t|x_0)]$ and propose several options for $q(x_0)$: 1) uniform noise 2) data generated by the learned model 3) result of MCMC sampling. 


\vspace{-3mm}
\subsection{Self normalized TCSIS}\label{sec:TCSIS_snis}
\vspace{-2mm}



Here we propose a practical algoritm for learning the concrete score. One can directly utilize the Proposition~\ref{th:TCS_Identity} and estimate both numerator and denominator by Monte Carlo.

\vspace{-4mm}
\begin{equation}\label{eq:MC_concrete_score}
    s_{t}(x)_y \approx \hat{s}_{t}(x)_y = \frac{ \sum_{i=0}^N \overline{p}(x_0^{i, y})}{\sum_{i=0}^N \overline{p}(x_0^{i, x})}, \quad x_0^{i, y} \sim p_{t|0}(x_0|y), x_0^{i, x} \sim p_{t|0}(x_0|x)
\end{equation}
\vspace{-3mm}

Then one can derive learning objective


\vspace{-5mm}
\hspace{-3mm}
\begin{equation}\label{eq:loss_SNIS}
    \hat{L}^{\text{SNIS}}_\theta = \mathbb{E}_{q(x_t), t}[\sum_y D^{\text{SE}}(\hat{s}_{t}(x_t)_y, s_{t}^\theta(x_t)_y)] = \mathbb{E}_{q(x_t), t}[\sum_y(s_{t}^\theta(x_t)_y - \hat{s}_{t}(x_t)_y \log s_{t}^\theta(x_t)_y)] + C,
\end{equation}
\vspace{-5mm}

where we have used Score Entropy \cite{lou2024discrete} in (\ref{eq:TCSIS_base}) as $D$, but generally any $D$ can be used. This formulation enables the parametrization of the concrete score $s_{t}^\theta(x_t)_y$ in log-space, i.e., $\log s_{t}(x_t)_y \approx \log s_{t}^\theta(x_t)_y$,  which improves the stability of the method. In addition, similar to Eq. 9 in \cite{akhound2024iterated_iDEM} one can formulate  Eq~\ref{eq:MC_concrete_score} as a self normalized importance sampling, see Appendix~\ref{appx:proofs}
We call sampler parametrized and learned in such way \textit{Self Normalized TCSIS}. 



\vspace{-3mm}
\subsection{Unbiased TCSIS}\label{sec:TCSIS_unbiased}
\vspace{-2mm}

On the other hand it is possible to parametrize by neural network not the concrete score, but the unnormalized marginal density and construct concrete score out of it. Considering this parametrization the following proposition holds.


\begin{proposition}[Marginal densities parametrization]\label{pr:TCSIS_unbiased}
If $p_t^\theta(y)$ minimizes:

\vspace{-5mm}
\begin{equation}
    L^{p_t}_\theta = \mathbb{E}_{q(x_t), t}[D^{\text{SE}}(\overline{p}_t(x_t), p_{t}^\theta(x_t))] = \mathbb{E}_{q(x_t), t}[p_{t}^\theta(x_t) - \overline{p}_t(x_t) \log p_{t}^\theta(x_t)] + C, 
\end{equation}
where $\overline{p}_t(x) = \mathbb{E}_{p_{t|0}(x_0|x)}[\overline{p}(x)]$. Then the concrete score estimator constructed in the following way:  
    \vspace{-2mm}
    \begin{equation}
        s_t^\theta(x)_y = \frac{p_t^\theta(y)}{p_t^\theta(x)},
    \end{equation}
        
    
    \vspace{-3mm}
    minimizes (\ref{eq:TCSIS_base}). Furthermore, the estimation of objective $\hat{L}^{p_t}_\theta \approx L^{p_t}_\theta$, when $\overline{p}_t(x)$ is calculated via Monte Carlo, is \textit{unbiased}.

\vspace{-9mm}
\begin{equation}\label{eq:loss_unbiased}
    \hat{L}^{p_t}_\theta = \mathbb{E}_{q(x_t), t}[p_{t}^\theta(x_t) - \hat{\overline{p}}_t(x_t) \log p_{t}^\theta(x_t)] + C, \quad \hat{\overline{p}}_t(x) = \sum_{i}^N \overline{p}(x_0^i), \quad x_0^i \sim p_{t|0}(x_0|x)
\end{equation}

\end{proposition}
\vspace{-3mm}




We call sampler parametrized and learned in such way \textit{Unbiased TCSIS}. The most important thing about Proposition~\ref{pr:TCSIS_unbiased} is unbiasedness of the leaning objective, which is valuable. We note that the choice of using here the Score Entropy as D in Eq~\ref{eq:TCSIS_base} is crucial as it allows for unbiased estimation with parametrization of $p_t^\theta(x)$ is log space. The neural network parametrization $\log p^\theta_t: \mathcal{X}^d \times \mathbb{R} \rightarrow \mathbb{R}$ allows to assemble the concrete score in log space during the inference time, i.e., $\log s^\theta_t(x)_y = \log p^\theta_t(y) - \log p^\theta_t(x)$, however under the factorization assumption it  requires $\mathcal{O}(dV)$ neural network evaluations.


\begin{figure}[!t]
    \centering
    \vspace{-21mm}
    \begin{subfigure}[b]{0.35\textwidth}
        \includegraphics[width=\linewidth]{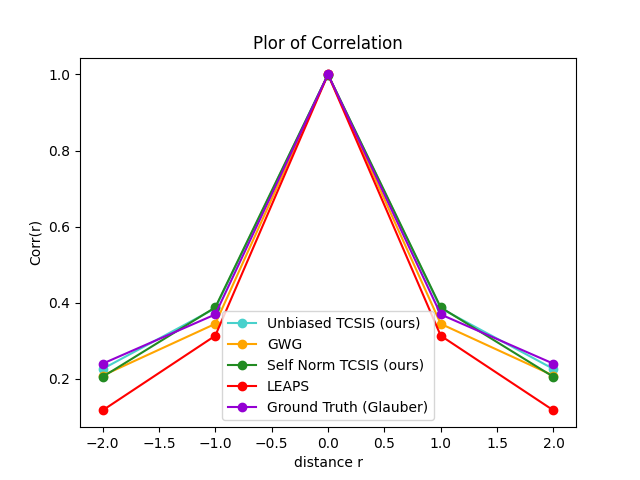}
        \caption{$\beta_{\rm high}$}
        \label{fig:sub1}
    \end{subfigure}
    \hspace{-6mm}
    \begin{subfigure}[b]{0.35\textwidth}
        \includegraphics[width=\linewidth]{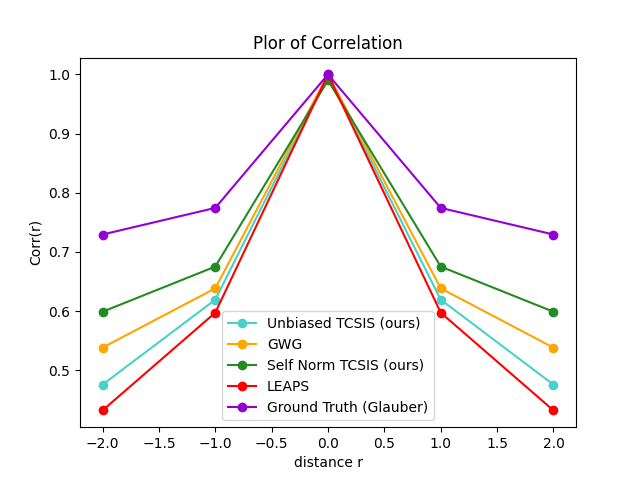}
        \caption{$\beta_{\rm critical}$}
        \label{fig:sub2}
    \end{subfigure}
    \hspace{-6mm}
    \begin{subfigure}[b]{0.35\textwidth}
        \includegraphics[width=\linewidth]{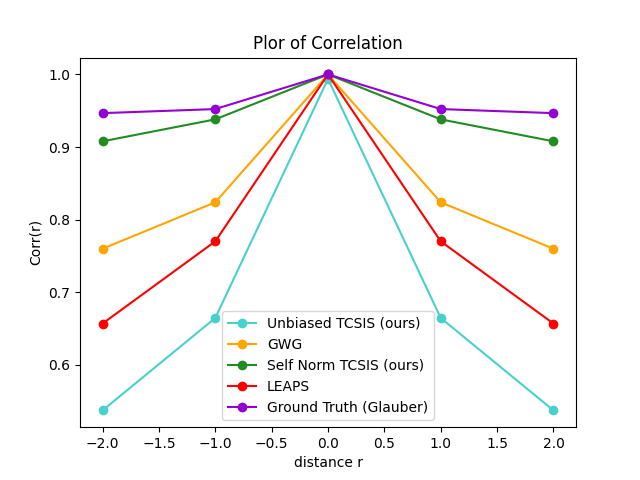}
        \caption{$\beta_{\rm low}$}
        \label{fig:sub3}
    \end{subfigure}
    
    \caption{Ising $L=4$ lattice model correlation plots. The closer to ground truth the better.}
    \label{fig:three_subfigures}
\end{figure}

\begin{figure}[!t]
    \centering
    \vspace{-5mm}
    \begin{subfigure}[b]{0.35\textwidth}
        \includegraphics[width=\linewidth]{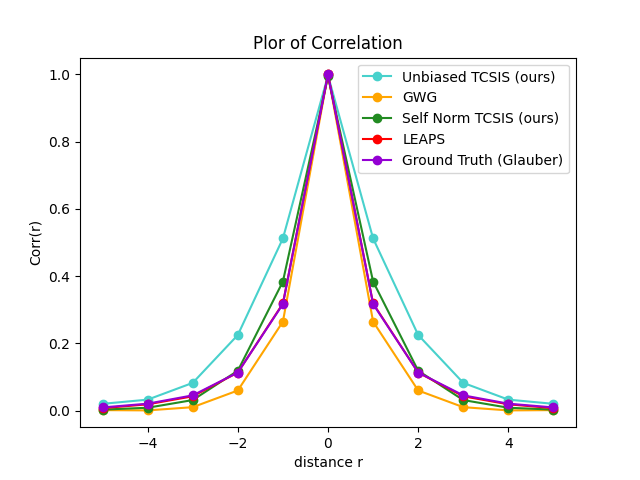}
        \caption{$\beta_{\rm high}$}
        \label{fig:sub1}
    \end{subfigure}
    \begin{subfigure}[b]{0.35\textwidth}
        \includegraphics[width=\linewidth]{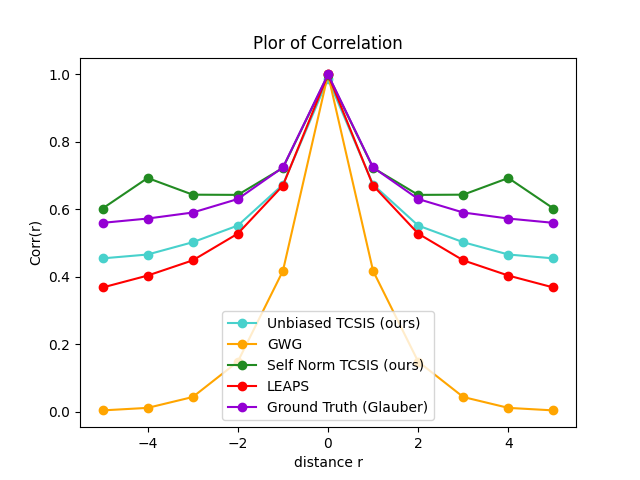}
        \caption{$\beta_{\rm critical}$}
        \label{fig:sub2}
    \end{subfigure}
       
    \caption{Ising $L=10$ lattice model correlation plots. The closer to ground truth the better.}
    \label{fig:three_subfigures}
    \vspace{-6mm}
\end{figure}

\vspace{-5mm}
\section{Experiments}
\vspace{-3mm}

We evaluate both our methods: Self Normalized TCSIS and Unbiased TCSIS on problem set from statistical physics, i.e., the lattice Ising model. The Ising model on $L \times L$ lattice (with no magnetization) is defined by its energy distribution: 
$p(x) \propto \exp(\beta \sum_{\langle i,  j\rangle}x_ix_j),$
where $x \in \{x_i\}_{i=1}^d$, $x_i \in \{0, 1\}$, $\beta$ is the inverse temperature (interaction strength) and $\sum_{<i, j>}$ denotes the summation across nearest neighboring spins $x_i, x_j$ on the lattice. Neighboring spins are uncorrelated at high temperature but reach a critical correlation when the temperature drops behold a certain threshold. 

We test our model on Ising lattices with inverse temperatures $\beta_{\text{high}}=0.28$,  $\beta_{\text{critical}}=0.4407$, $\beta_{\text{low}}=0.6$, since Ising model behaviour is known to be distinct at these temperatures by the phase transition theory \cite{kramers1941statistics_ising}. In particular, we run on $L=4$ lattice with all of the temperatures and on $L=10$ lattice with $\beta_{\text{high}}, \beta_{\text{critical}}$. To validate our model we compute two point correlations, see Appendix~\ref{appx:two_point_corr} and compare with another neural sampler LEAPS \cite{holderriethleaps} and MCMC method GWG \cite{grathwohl2021oops_GWG}, all with $24$ inference steps on $L=4$ lattice and  $64$ inference steps on $L=10$ lattice. The LEAPS sampler is run without importance sampling correction. As a ground truth we run Glauber dynamics for 10000 steps. For description of metrics, all the methods hyperparameters, experimental details and additional results, see Appendix~\ref{appx:exp}.





We find that Self-Normalized TCSIS consistently performs well across all temperature regimes and experimental settings, outperforming LEAPS. The Unbiased TCSIS achieves performance comparable to LEAPS overall. While the MCMC baseline performs strongly on small $L=4$ lattices, its accuracy deteriorates as dimensionality increases and temperature decreases, which is a known challenge for MCMC methods \cite{DISCS_2023}. In contrast, neural samplers demonstrate better scalability and maintain competitive performance in these regimes.

\vspace{-5mm}
\section{Discussion}
\vspace{-3mm}
We introduce the Target Concrete Score Identity and the corresponding family of Target Concrete Score Identity Samplers with simulation free training. Empirical results demonstrate that our approach is competitive across the considered tasks. Moreover, the framework of employing the concrete score for sampling establishes a promising direction for advancing the development of more efficient methods beyond our current approach. 


Future improvements include approximating the concrete score via Taylor expansion \cite{ou2025dnfs, grathwohl2021oops_GWG}, exploring alternative architectures for Unbiased TCSIS to reduce the $\mathcal{O}(dV)$ inference cost, and developing a learning-free variant where the concrete score is estimated online using Proposition~\ref{th:TCS_Identity}.

\paragraph{Limitations}\label{appx:limitations}

We find both TCSIS methods struggle with growth of problem dimensionality and at lower temperatures. Self Normalized TCSIS requires a lot of energy evaluations for training, i.e., $\mathcal{O}(dVN)$ calls of energy function, which makes it inefficient for environments with costly calls of energy function, e.g., neural network. However worth noting that Taylor approximation and Top-k sampling can help to alleviate the problem \cite{zhang_target_thornton, grathwohl2021oops_GWG}.

Unbiased TCSIS experiences costly inference stage, because one has to assemble the concrete score from approximated marginal densities, which requires $\mathcal{O}(dV)$ calls of $p_{t}^\theta(x)$ neural network. The Unbiased TCSIS loss $\hat{L}^{p_t}$ (\ref{eq:loss_unbiased}) has the $p^\theta_t(x_t)$ and $\hat{\overline{p}}_t(x_t)$ terms that do experience exponential asymptotic and can induce overflow in the values of energy function.

\bibliographystyle{plain}
\bibliography{references}

\begin{thebibliography}{10}

\bibitem{kramers1941statistics_ising}
Statistics of the two-dimensional ferromagnet. part i.
\newblock {\em Physical Review}, 60(3):252, 1941.

\bibitem{akhound2024iterated_iDEM}
Tara Akhound-Sadegh, Jarrid Rector-Brooks, Joey Bose, Sarthak Mittal, Pablo Lemos, Cheng-Hao Liu, Marcin Sendera, Siamak Ravanbakhsh, Gauthier Gidel, Yoshua Bengio, et~al.
\newblock Iterated denoising energy matching for sampling from boltzmann densities.
\newblock In {\em International Conference on Machine Learning}, pages 760--786. PMLR, 2024.

\bibitem{austin2021structured}
Jacob Austin, Daniel~D Johnson, Jonathan Ho, Daniel Tarlow, and Rianne Van Den~Berg.
\newblock Structured denoising diffusion models in discrete state-spaces.
\newblock {\em Advances in neural information processing systems}, 34:17981--17993, 2021.

\bibitem{campbell2022continuous}
Andrew Campbell, Joe Benton, Valentin De~Bortoli, Thomas Rainforth, George Deligiannidis, and Arnaud Doucet.
\newblock A continuous time framework for discrete denoising models.
\newblock {\em Advances in Neural Information Processing Systems}, 35:28266--28279, 2022.

\bibitem{de2024target_TSM}
Valentin De~Bortoli, Michael Hutchinson, Peter Wirnsberger, and Arnaud Doucet.
\newblock Target score matching.
\newblock {\em arXiv preprint arXiv:2402.08667}, 2024.

\bibitem{DISCS_2023}
Katayoon Goshvadi, Haoran Sun, Xingchao Liu, Azade Nova, Ruqi Zhang, Will Grathwohl, Dale Schuurmans, and Hanjun Dai.
\newblock Discs: A benchmark for discrete sampling.
\newblock In A.~Oh, T.~Naumann, A.~Globerson, K.~Saenko, M.~Hardt, and S.~Levine, editors, {\em Advances in Neural Information Processing Systems}, volume~36, pages 79035--79066. Curran Associates, Inc., 2023.

\bibitem{grathwohl2021oops_GWG}
Will Grathwohl, Kevin Swersky, Milad Hashemi, David Duvenaud, and Chris Maddison.
\newblock Oops i took a gradient: Scalable sampling for discrete distributions.
\newblock In {\em International Conference on Machine Learning}, pages 3831--3841. PMLR, 2021.

\bibitem{havensadjoint}
Aaron~J Havens, Benjamin~Kurt Miller, Bing Yan, Carles Domingo-Enrich, Anuroop Sriram, Daniel~S Levine, Brandon~M Wood, Bin Hu, Brandon Amos, Brian Karrer, et~al.
\newblock Adjoint sampling: Highly scalable diffusion samplers via adjoint matching.
\newblock In {\em Forty-second International Conference on Machine Learning}.

\bibitem{heo2024rotary}
Byeongho Heo, Song Park, Dongyoon Han, and Sangdoo Yun.
\newblock Rotary position embedding for vision transformer.
\newblock In {\em European Conference on Computer Vision}, pages 289--305. Springer, 2024.

\bibitem{holderriethleaps}
Peter Holderrieth, Michael~Samuel Albergo, and Tommi Jaakkola.
\newblock Leaps: A discrete neural sampler via locally equivariant networks.
\newblock In {\em Forty-second International Conference on Machine Learning}.

\bibitem{lartillot2004bayesian}
Nicolas Lartillot and Herv{\'e} Philippe.
\newblock A bayesian mixture model for across-site heterogeneities in the amino-acid replacement process.
\newblock {\em Molecular biology and evolution}, 21(6):1095--1109, 2004.

\bibitem{lireheated_DMALA}
Muheng Li and Ruqi Zhang.
\newblock Reheated gradient-based discrete sampling for combinatorial optimization.
\newblock {\em Transactions on Machine Learning Research}.

\bibitem{lou2024discrete}
Aaron Lou, Chenlin Meng, and Stefano Ermon.
\newblock Discrete diffusion modeling by estimating the ratios of the data distribution.
\newblock In {\em Proceedings of the 41st International Conference on Machine Learning}, pages 32819--32848, 2024.

\bibitem{meng2022concrete}
Chenlin Meng, Kristy Choi, Jiaming Song, and Stefano Ermon.
\newblock Concrete score matching: Generalized score matching for discrete data.
\newblock {\em Advances in Neural Information Processing Systems}, 35:34532--34545, 2022.

\bibitem{metropolis1953equation}
Nicholas Metropolis, Arianna~W Rosenbluth, Marshall~N Rosenbluth, Augusta~H Teller, and Edward Teller.
\newblock Equation of state calculations by fast computing machines.
\newblock {\em The journal of chemical physics}, 21(6):1087--1092, 1953.

\bibitem{murray2006mcmc}
Iain Murray, Zoubin Ghahramani, and David~JC MacKay.
\newblock Mcmc for doubly-intractable distributions.
\newblock In {\em Proceedings of the Twenty-Second Conference on Uncertainty in Artificial Intelligence}, pages 359--366, 2006.

\bibitem{neal2011mcmc}
Radford~M Neal et~al.
\newblock Mcmc using hamiltonian dynamics.
\newblock {\em Handbook of markov chain monte carlo}, 2(11):2, 2011.

\bibitem{newman1999monte}
MEJ Newman and GT~Barkema.
\newblock Monte carlo methods in statistical physics.
\newblock {\em Monte Carlo methods in statistical physics/MEJ Newman and GT Barkema. Oxford: Clarendon Press}, 1999.

\bibitem{ou2025dnfs}
Zijing Ou, Ruixiang Zhang, and Yingzhen Li.
\newblock Discrete neural flow samplers with locally equivariant transformer.
\newblock {\em arXiv preprint arXiv:2505.17741}, 2025.

\bibitem{peebles2023scalable_dit}
William Peebles and Saining Xie.
\newblock Scalable diffusion models with transformers.
\newblock In {\em Proceedings of the IEEE/CVF international conference on computer vision}, pages 4195--4205, 2023.

\bibitem{sahoo2024simple_MDLM}
Subham Sahoo, Marianne Arriola, Yair Schiff, Aaron Gokaslan, Edgar Marroquin, Justin Chiu, Alexander Rush, and Volodymyr Kuleshov.
\newblock Simple and effective masked diffusion language models.
\newblock {\em Advances in Neural Information Processing Systems}, 37:130136--130184, 2024.

\bibitem{sanokowskiscalable_SDDS}
Sebastian Sanokowski, Wilhelm~Franz Berghammer, Haoyu~Peter Wang, Martin Ennemoser, Sepp Hochreiter, and Sebastian Lehner.
\newblock Scalable discrete diffusion samplers: Combinatorial optimization and statistical physics.
\newblock In {\em The Thirteenth International Conference on Learning Representations}.

\bibitem{schiffsimple_UDLM}
Yair Schiff, Subham~Sekhar Sahoo, Hao Phung, Guanghan Wang, Sam Boshar, Hugo Dalla-torre, Bernardo~P de~Almeida, Alexander~M Rush, Thomas PIERROT, and Volodymyr Kuleshov.
\newblock Simple guidance mechanisms for discrete diffusion models.
\newblock In {\em The Thirteenth International Conference on Learning Representations}.

\bibitem{vargasdenoising}
Francisco Vargas, Will~Sussman Grathwohl, and Arnaud Doucet.
\newblock Denoising diffusion samplers.
\newblock In {\em The Eleventh International Conference on Learning Representations}.

\bibitem{vargastransport_CMCD}
Francisco Vargas, Shreyas Padhy, Denis Blessing, and Nikolas N{\"u}sken.
\newblock Transport meets variational inference: Controlled monte carlo diffusions.
\newblock In {\em The Twelfth International Conference on Learning Representations}.

\bibitem{zhangpath_PIS}
Qinsheng Zhang and Yongxin Chen.
\newblock Path integral sampler: A stochastic control approach for sampling.
\newblock In {\em International Conference on Learning Representations}.

\bibitem{zhang_target_thornton}
Ruixiang ZHANG, Shuangfei Zhai, Yizhe Zhang, James Thornton, Zijing Ou, Joshua~M Susskind, and Navdeep Jaitly.
\newblock Target concrete score matching: A holistic framework for discrete diffusion.
\newblock In {\em Forty-second International Conference on Machine Learning}.

\end{thebibliography}

\newpage

\appendix





\section{Proofs} \label{appx:proofs}


\subsection{Proof of Target Concrete Score Identity}




\begin{proof}
    $$p_t(x_t)= \mathbb{E}_{p(x_0)}[p_{t|0}(x_t|x_0)]= \sum_{x_0}p_{t|0}(x_t|x_0)p(x_0) = \big[\text{Eq~\ref{eq:uniform_forward}}\big] = \sum_{x_0}\prod_{i=0}^d p_{t|0}(x_t^i|x_0^i)p(x_0)$$

    Then let us analyze $p_{t|0}(x_t^i|x_0^i)$:

    $$p_{t|0}(x_t^i|x_0^i) = \exp(\int_{0}^t\sigma_sds Q^{\rm  uniform}(x_t^i, x_0^i)),$$

    Next, recap that $Q^{\rm uniform}$ has the form:

        $$Q^{\rm uniform} = \begin{bmatrix} 1 - N & 1 & \cdots & 1\\ 1 & 1 - N & \cdots & 1\\ \vdots & \vdots & \ddots & \vdots \\ 1 & 1 & \cdots & 1 - N\end{bmatrix}$$

    and is \textit{symmetric} by construction. Note that an exponent of symmetric matrix is as well symmetric. Then it follows:

    $$p_{t|0}(x_t^i|x_0^i) = \exp(\int_{0}^t\sigma_sds Q^{\rm  uniform}(x_t^i, x_0^i)) = \big[ {\rm symmetric} \big] = \exp(\int_{0}^t\sigma_sds Q^{\rm  uniform}(x_t^i, x_0^i)) = p_{t|0}(x_0^i|x_t^i).$$

    Getting back to marginal density:

    $$p_t(x_t)=\sum_{x_0}\prod_{i=0}^d p_{t|0}(x_t^i|x_0^i)p(x_0) = \sum_{x_0}\prod_{i=0}^d p_{t|0}(x_0^i|x_t^i)p(x_0) = $$
    $$= \sum_{x_0}p_{t|0}(x_0|x_t)p(x_0)  = \mathbb{E}_{p_{t|0}(x_0|x_t)}[p(x_0)]$$

    Then one can construct a concrete score with unnormarlized density $p(x_0) = \frac{\overline{p}(x_0)}{Z}$:

    $s_t(x)_y = \frac{p_t(y)}{p_t(x)} = \frac{\mathbb{E}_{p_{t|0}(x_0|y)}[p(x_0)]}{\mathbb{E}_{p_{t|0}(x_0|x)}[p(x_0)]} = \big[ p(x_0) = \frac{\overline{p}(x_0)}{Z} \big] = \frac{\mathbb{E}_{p_{t|0}(x_0|y)}[\frac{p(x_0)}{Z}]}{\mathbb{E}_{p_{t|0}(x_0|x)}[\frac{p(x_0)}{Z}]} = \frac{\mathbb{E}_{p_{t|0}(x_0|y)}[\overline{p}(x_0)]}{\mathbb{E}_{p_{t|0}(x_0|x)}[\overline{p}(x_0)]}$
    
\end{proof}
\subsection{Self normalized TCSIS}

One can show that objective \ref{eq:loss_SNIS} can be interpreted as self normalized importance weighted estimate,

\begin{gather}
    s_t(x)_y = \frac{\mathbb{E}_{p_{t|0}(x_0|y)}[\overline{p}(x_0)]}{\mathbb{E}_{p_{t|0}(x_0|x)}[\overline{p}(x_0)]} = \frac{\sum_{x_0 \in S} p_{t|0}(x_0|y) \overline{p}(x_0)}{ \sum_{x_0 \in S} p_{t|0}(x_0|x)\overline{p}(x_0)} =\\
    \frac{\sum_{x \in S} \frac{p_{t|0}(x_0|y)}{p_{t|0}(x_0|x)}p_{t|0}(x_0|x) \overline{p}(x_0)}{ \sum_{x_0 \in S} p_{t|0}(x_0|x)\overline{p}(x_0)} =   \frac{\mathbb{E}_{p_{t|0}(x_0|x)}\big[\frac{p_{t|0}(x_0|y)}{p_{t|0}(x_0|x)}\overline{p}(x_0)\big]}{\mathbb{E}_{p_{t|0}(x_0|x)}[\overline{p}(x_0)]},
\end{gather}

which can be interpreted as expectation of $\frac{p_{t|0}(x_0|y)}{p_{t|0}(x_0|x)}$ under $p_{t|0}(x_0|x)$ weighted by $w_i \approx \frac{\overline{p}(x_0^i)}{\sum_i \overline{p}(x_0^i)}$. 

\section{Algorithm}



Here we present an algorithm for training both the Self Normalized TCSIS and Unbiased TCSIS, see Algorithm~\ref{alg:TCSIS_train}. 

\begin{algorithm}[t!]
    \caption{Training of TCSIS.}
    \label{alg:TCSIS_train}
    \SetKwInOut{Input}{Input}\SetKwInOut{Output}{Output}
    \Input{Energy function $E(x)$, Neural network parameters $\theta$ initialization, $T$ number of grad steps, "Proposal" distribution $q(x_t)$, Bacth Size $L$, Number of MC samples $L$, TCSIS Type}
    
    \Output{Neural network parameters $\theta$}
    \For{$i \gets 1$ \KwTo $T$}{
    Sample batch $\{x_t^m\}_{m=1}^M \sim q(x_t)$; \\
    Get all the 1-Hamming distance sequences $\{y_t^{m, k, n}\}_{n=1, k=1}^{M, V, N}$ for $\{x_t^m\}_{m=1}^M$; \\
    Sample $\{\hat{x}_t^{m, k, n, l}\}_{m=1, k=1, n=1, l=1}^{M, V, N, L} \sim p_{t|0}(x_0|\{y_t^{m, k, n}\})$; \\
    $T^E \leftarrow E(\{\hat{x}_t^{m, k, n, l}\}_{m=1, k=1, n=1, l=1}^{M, V, N, L})$;\\
    $T^{\rm noise} \leftarrow \log p_{t|0}(\{\hat{x}_t^{m, k, n, l}\}_{m=1, k=1, n=1, l=1}^{M, V, N, L}|\{y_t^{m, k, n}\})$ ; \\
    Compute marginal density estimation $\hat{\overline{p}}_t(x_t) = T^E + T^{\rm noise}$;\\
    \eIf{TCSIS type = Self-Normalized}{
    Compute concrete score estimation $\hat{s}_t(x)_y$ using Eq~\ref{eq:MC_concrete_score} and $\hat{\overline{p}}_t(x_t)$;\\
    Compute loss $L_\theta$ using Eq~\ref{eq:loss_SNIS} and $\hat{s}_t(x)_y$;\\
    }{
    Compute loss $L_\theta$ using Eq~\ref{eq:loss_unbiased} and $\hat{\overline{p}}_t(x_t)$;;\\}
    Optimize $\theta \leftarrow \theta - \nabla_\theta L_\theta$;\\
    }
\end{algorithm}


\section{Additional related works}

\subsection{Uniform discrete diffusion models}\label{appx:uniform_diff}

In uniform diffusion models the forward noising kernel is defined by factorized rate matrix $Q_t(x^i, y^i)=\sigma_tQ^{\rm uniform}(x^i, y^i)$, where:

        $$Q^{\rm uniform} = \begin{bmatrix} 1 - N & 1 & \cdots & 1\\ 1 & 1 - N & \cdots & 1\\ \vdots & \vdots & \ddots & \vdots \\ 1 & 1 & \cdots & 1 - N\end{bmatrix}$$

The marginal densities ODE, i.e., Eq~\ref{eq:CTMC_Kolmogorov_forward}, can be integrated analytically and transition probability $p_{t|0}(x_t|x_0)$ would take the form:

\begin{equation}\label{eq:uniform_forward}
p_{t|0}(x_t|x_0) = \prod_{i=0}^N p_{t|0}(x_t^i|x_0^i) = \prod_{i=0}^N \exp(\overline{\sigma}_tQ^{\rm uniform}(x_t^i, x_0^i)), \quad Q^{\text{uniform}}(x, y) = \left\{
    \begin{array}{l}
    1 , y \neq x \\
    1 - N, y=x
    \end{array}
    \right.
\end{equation}

where $x_t^i$ is $i$-th element of $x$. With sufficiently large time $T$ forward process converges to \textbf{uniform noise}.


\subsection{Factorization and concrete score for 1-Hamming distance neighborhood.} \label{appx:factorization}


In practice, concrete score is learned only for sequences withing 1-Hamming neighborhood Section 3.3 \cite{lou2024discrete}, i.e., between sequences that do differ at most by one token. Modeling concrete score between general sequences $\mathcal{X}=\{1,...,N\}^d$ in computationally expensive as number of states grow exponentially. In addition, modeling of concrete scores withing 1-Hamming neighborhood, i.e., factorized by sequence length, is sufficient for consistent modeling of CTMC processes \cite{campbell2022continuous}. 

Modeling of concrete scores only for sequences withing 1-Hamming neighborhood is equivalent to modeling factorized rate matrices $Q(x, y)$ or factorized probability distributions $p(x)= \prod_{i=0}^dp(x^i)$ \cite{lou2024discrete, campbell2022continuous}. In that case factorized concrete score $s_t(x): \mathcal{X}^d \times \mathbb{R} \rightarrow \mathbb{R}^{N \times d}$ would be represented by a \textit{matrix}, rather then the full concrete score $s_t(x): \mathcal{X}^d \times \mathbb{R} \rightarrow \mathbb{R}^{|\mathcal{X}^d|}$ that can be represented by the \textit{vector of exponential size} $|\mathcal{X}^d|$.



\section{Experimental details}\label{appx:exp}

\subsection{Magnetization}

Here we present the magnetization histograms for our experiments. 
The magnetization is the sum of all coordinate values on a lattice computed as follows: $M(x)=\sum_ix_i$. The histograms of magnetizations for Ising models on $L=4$ and $L=10$ lattices are presented in Figures \ref{fig:mag_hist_l_4} and \ref{fig:mag_hist_l_4} correspondingly. Magnetization histogramms for each method were calculated using $10000$ samples.

\begin{figure}[!t]
    \centering
    \begin{subfigure}[b]{0.35\textwidth}
        \includegraphics[width=\linewidth]{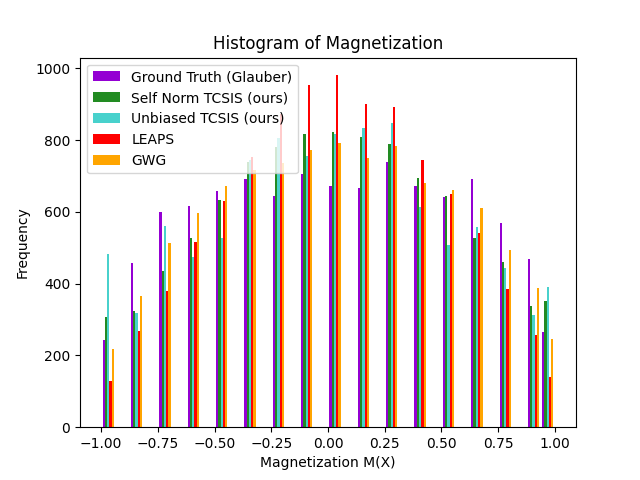}
        \caption{High temperature setting}
        \label{fig:sub1}
    \end{subfigure}
    \hspace{-6mm}
    \begin{subfigure}[b]{0.35\textwidth}
        \includegraphics[width=\linewidth]{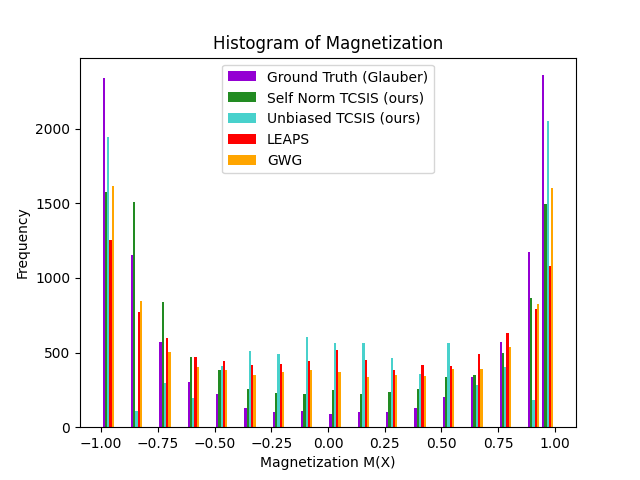}
        \caption{Critical temperature setting}
        \label{fig:sub2}
    \end{subfigure}
    \hspace{-6mm}
    \begin{subfigure}[b]{0.35\textwidth}
        \includegraphics[width=\linewidth]{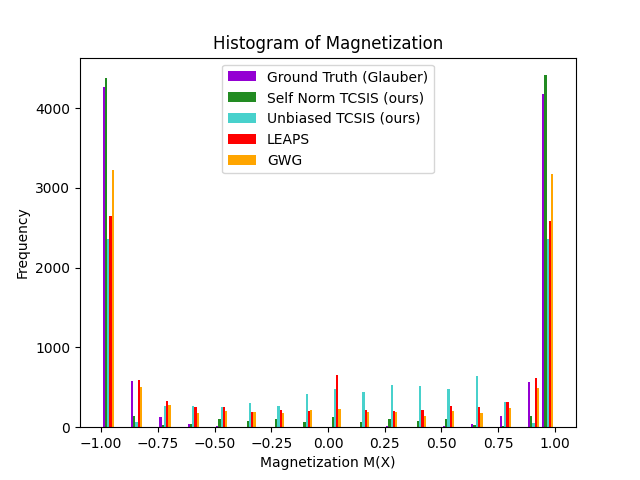}
        \caption{Low temperature setting}
        \label{fig:sub3}
    \end{subfigure}
    
    \caption{Ising $L=4$ lattice model with different inverse temperatures magnetization histograms.}
    \label{fig:mag_hist_l_4}
\end{figure}

\begin{figure}[!t]
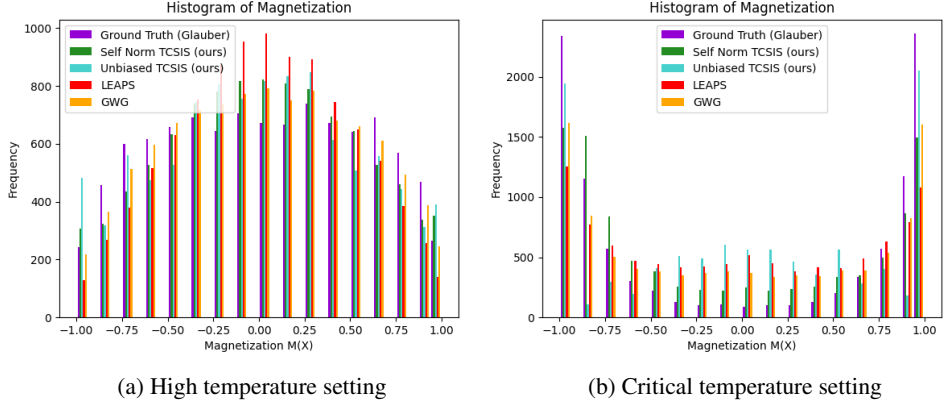

    \centering
    \begin{subfigure}[b]{0.49\textwidth}
        \includegraphics[width=\linewidth]{figures/L_4_temp_0.28_mag.png}
        \caption{High temperature setting}
        \label{fig:sub1}
    \end{subfigure}
    \hspace{-6mm}
    \begin{subfigure}[b]{0.49\textwidth}
        \includegraphics[width=\linewidth]{figures/L_4_temp_0.4407_mag.png}
        \caption{Critical temperature setting}
        \label{fig:sub2}
    \end{subfigure}
    
    \caption{Ising $L=10$ lattice model with different inverse temperatures magnetization histograms.}
    \label{fig:mag_hist_l_4}
\end{figure}

\subsection{Two point Correlation}\label{appx:two_point_corr}

The two point correlation on lattice is defined as follows:

$$G(r) = \mathbb{E}[x_ix_{i+r}] - \mathbb{E}[x_i]\mathbb{E}[x_{i+r}],$$

where $x_i$ and $x_{i+r}$ are separated by distance of $r$ on the lattice.  The latter is a measure of the dependency between spin values per distance $r$ lattice separation.

\subsection{TCSIS hyperparameters}

The main part of the hyperpameters is described in the Table~\ref{tab:hyperparams}. Separately, as a neural network we use diffusion transformer (DiT) \cite{peebles2023scalable_dit} with 2-dimensional rotary embeddings \cite{heo2024rotary} and for Unbiased TCSIS parametrization we add a regression head on top. In all the cases neural network has fewer then 1M parameters. In "proposal" $q(x_t)$ column: Noise - random uniform noise, Model and Noise - a mixture of samples from model and random noise (with weight of $0.9$ for model samples), GWG for samples from GWG \cite{grathwohl2021oops_GWG} MCMC method with 1000 steps. For Model and GWG samples replay buffer was used. Precond means preconditioning that was used for the Unbiased TCSIS. Following \cite{zhangpath_PIS} we model marginal densities as $\log p_t^\theta(x) = NN_\theta(x, t) + NN_\theta(t) * E(x)$, where $E(x)$ is the energy function of corresponding Ising model in our case. Learning of Unbiased TCSIS in $L=10$ Ising model setting was carried out in \underline{$torch.double$} because the loss values could induce overflow. We use a 0.999 EMA in all the TCSIS setups. Training of TCSIS took less then 6 hours on NVIDIA A100 GPU and inference took less than 5 minutes on NVIDIA A100 GPU.

\begin{table}[h!]
\centering
\begin{tabular}{|l|c|c|c|c|c|c|c|}
\hline
\textbf{Method} & Lattice dim, $L$ & MC N & "Proposal" $q(x_t)$ & LR & Grad steps & Batch size & Precond \\
\hline
Self Norm TCSIS & 4 & 500 & Noise & $1e-5$ & 50k & 64 & \ding{55}\\
Unbiased TCSIS & 4 & 100 & GWG & $1e-4$ & 50k & 64 & \ding{51}\\
Self Norm TCSIS & 10 & 2000 & Model and Noise & $3e-6$ & 150k & 64 & \ding{55}\\
Unbiased TCSIS & 10 & 500 & GWG & $1e-4$ & 100k & 64 & \ding{51}\\
\hline
\end{tabular}
\caption{Comparison of methods with their configurations.}
\label{tab:hyperparams}
\end{table}
\vspace{-8mm}

\subsection{LEAPS}

LEAPS \cite{holderriethleaps} official github implementation was taken:

\begin{center}
\url{https://github.com/malbergo/leaps}
\end{center}

Locally-equivariant convolutional (LEC) network was taken as a neural network with kernel sizes $[3, 3, 3, 3]$ for $L=4$ and $[3, 5, 7, 9]$ for $L=10$, 20 channels were taken for each setup. MLP with $[200, 200, 200, 200]$ channels was taken as a free energy network. $20000$ geometric annealing gradient steps were taken as a warmup. All the other hyperparameters were taken as default from the proposed by authors configuration. LEAPS was ran without correction during sampling. LEAPS training took less than 24 hours on NVIDIA A100 GPU and inference took less than 5 minutes on NVIDIA A100 GPU.

\subsection{GWG}

The implementation was taken from the DMALA \cite{lireheated_DMALA} repository:

\begin{center}
\url{https://github.com/ruqizhang/discrete-langevin}
\end{center}

We used a Taylor approximation for the difference function, disallowed multiple jumps per step, and ran the $L=4$ setup with 24 transition steps and $L=10$ setup with 64 transition steps, a randomized proposal, temperature set to 2, and step size of 0.05. Inference of the MCMC took less than 1 minute on NVIDIA A100 GPU.




\end{document}